\documentclass[letterpaper, 10 pt, conference]{ieeeconf}  

\usepackage{cite}
\usepackage{amsmath,amssymb,amsfonts}
\usepackage{algorithmic}
\usepackage{graphicx}
\usepackage{textcomp}
\usepackage{xcolor}
\usepackage[labelfont=bf,font=small]{caption}
\usepackage{hyperref}

\usepackage{booktabs}

\IEEEoverridecommandlockouts                              

\title{\LARGE \bf
Examining Audio Communication Mechanisms for \\ Supervising Fleets of Agricultural Robots}

\author{Abhi Kamboj, Tianchen Ji, and Katie Driggs-Campbell
\thanks{A. Kamboj, T. Ji, and K. Driggs-Campbell are with the department of Electrical and Computer Engineering at the University of Illinois, Urbana-Champaign
        {\tt\small \{akamboj2, tj12, krdc\}@illinois.edu}}%
\thanks{This work was supported by the USDA National Institute of Food and Agriculture (USDA/NIFA), through the National Robotics Initiative 2.0 (NIFA\#2021-67021-33449), the AI Institute AIFARMS through the Agriculture and Food Research Initiative (AFRI) (USDA/NIFA Award no. 2020-67021-32799), as well as the Illinois Center for Digital Agriculture.}%
}

\begin{document}

\maketitle
\thispagestyle{empty}
\pagestyle{empty}

\begin{abstract}

Agriculture is facing a labor crisis, leading to increased interest in fleets of small, under-canopy robots (agbots) that can perform precise, targeted actions (e.g., crop scouting, weeding, fertilization), while being supervised by human operators remotely. 
However, farmers are not necessarily experts in robotics technology and will not adopt technologies that add to their workload or do not provide an immediate payoff.
In this work, we explore methods for communication between a remote human operator and multiple agbots and examine the impact of audio communication on the operator's preferences and productivity.
We develop a simulation platform where agbots are deployed across a field, randomly encounter failures, and call for help from the operator. 
As the agbots report errors, various audio communication mechanisms are tested to convey which robot failed and what type of failure occurs. 
The human is tasked with verbally diagnosing the failure while completing a secondary task. 
A user study was conducted to test three audio communication methods: earcons, single-phrase commands, and full sentence communication. 
Each participant completed a survey to determine their preferences and each method's overall effectiveness. 
Our results suggest that the system using single phrases is the most positively perceived by participants and may allow for the human to complete the secondary task more efficiently. The code is available at: 
\url{https://github.com/akamboj2/Agbot-Sim}.
\end{abstract}

\section{Introduction}

Agriculture is currently facing a human labor crisis~\cite{CFBF2019}, harming profitability and causing negative downstream effects. 
As a result, precise actions (e.g., weeding, targeted pesticides) are not feasible at the scale required for annual row crops (e.g., corn, soybeans), which dominate the Midwest and much of the US.
In these settings, agriculture is only practical with heavy reliance on fertilizers, pesticides, and herbicides applied with large equipment (e.g., tractors, combines)~\cite{foley2011solutions,Godfray2010Food}. 
While this equipment is familiar to farmers and can be automated to alleviate labor concerns~\cite{ortiz2013evaluation}, such equipment is capital heavy, requires extra logistical oversight, introduces new safety risks to workers, and physically impacts the farm (e.g., soil-compaction, crop damage).

Small agricultural robots (agbots) can help alleviate the labor crisis and enable precision agriculture.
These agbots are designed to be small, inexpensive, and intelligent, and have seen growing attention in recent years~\cite{yaghoubi2013autonomous,pedersen2006agricultural}.  
To fully address the labor shortage, these agbots must be both easy to use and able to be deployed at scale, where one human is supervising many robots.
Despite growing interest in agbots, farmers already tend to be overwhelmed with the large number of equipment and data sources that are constantly made available to them~\cite{un-overload}.  
From large suppliers automating their products to new technologies for data collection~\cite{vasisht2017farmbeats}, farmers are being driven to manage a large set of equipment each with their own intricacies~\cite{CFBF2019}. 
Many human-robot interaction (HRI) methodologies focus on intuitive interface designs that allow for seamless integration of robots in a society of non-experts \cite{honig2018understanding,luria2019championing,robinson2021smooth, lupetti2021designerly}. 
Simulations are an effective method to test and overcome barriers preventing adoption of robots\cite{choi2021use,lemaignan2014simulation}. 
We follow a design focused simulation methodology to study which type of auditory interaction most positively influences a user’s perception and productivity in a remote monitoring setting. Our simulation stands in for actual agbots navigating fields of crops. The failure cases and solutions we simulate are identifiable and previously tested on a physical robot \cite{ji2020multi}.

We consider situations where a fleet of agbots are deployed in a field and a remote human operator supervises the robots.
As the robots navigate through crop rows, they randomly encounter failures that require help from the operator.
During each failure, the robot control center will audibly prompt the operator to provide assistance through verbal commands. 
Across experiments, we vary the audible prompt and measure the operator's perception of the system and productivity of completing a secondary task.

Handling failure scenarios is of critical importance for near-term deployment, as many agriculture tasks and environments are too complex for current agbots to handle without at least occasional failure~\cite{vasconez2019human,rodriguez2018beyond,ji2022proactive}. 
This setting follows the idea of sliding autonomy~\cite{dias2008sliding} and recent efforts to codify the levels of autonomy for field robots~\cite{LoA-field-robots}, which outlines how agbots with varying autonomous capabilities can interact with human operators~\cite{swamy2020scaled}. Failure cases have a strong effect on the user’s perception and trust in a system, meaning handling failures largely impacts the overall success of an HRI system~\cite{reig2021flailing}.

This research provides insight into the effectiveness and acceptance of audio communication interfaces when managing multiple autonomous robot failures. We studied the effect of earcons, single phrase commands, and full sentence speech on the user's perception of the system and their efficacy in completing a secondary task. 
We present three contributions: 
\begin{enumerate}
    \item We develop a simulated control center to explore how humans interact and monitor agricultural robots deployed across a field, while potentially encountering failures that require human assistance.
    \item {We demonstrate how audio signals (either tones or natural language)  improve an operator's efficiency and productivity compared to a traditional visual interface.}
    \item {Our user study provides insight on how well an operator perceives various auditory interaction systems in a remote robot monitoring setting, indicating which system will most effectively be adopted.} 
\end{enumerate}

This paper is organized as follows.
We review relevant literature in Section~\ref{sec:related-work}.
In Section~\ref{sec:methods}, we present an overview of our exploratory study, our hypotheses, and our measures.
Section~\ref{sec:results} and~\ref{sec:discussion} discuss the quantitative and qualitative findings from our user study.
Finally, we conclude in Section~\ref{sec:conclusion}.

\section{Related Work}
\label{sec:related-work}

\subsection{Audio in Design}
Auditory interfaces can be defined as bidirectional, communicative connections between two systems using audio, where audio refers to the production of sound\cite{peres2008auditory}. 
There are various methods in which sound can be used for auditory interaction in HRI. 
In HRI sound either deliberately communicates an intention, such as speech, notifications, or semantic-free utterances \cite{yilmazyildiz2016review}, or comes without intention such as consequential sound or movement sonification \cite{robinson2021smooth}. 
Using sound to intentionally convey information has many benefits including reducing visual overload, reinforcing visual messages, and providing additional information such as direction or emotion \cite{peres2008auditory}. 
Our study uses sound as a primary interface to reduce visual load for the human operator to complete a visually intensive secondary task. 
To the same degree that human-to-human correspondence involves various modes interaction such as audible, visual, and tactile, HRI also requires multi-modal interaction for successful integration of robots into society ~\cite{steil2003learning, kaber2006investigation}.  

Most studies about sound in HRI involve improving a user's perception of human-robot dialogue or nonverbal noises~\cite{marge2019miscommunication,robinson2021smooth}, ignoring the wide range of possibilities that auditory interaction can offer~\cite{frauenberger2007survey}. 
There are four main ways data can be encoded into audio: auditory icons, earcons, sonification and speech~\cite{mcgookin2004understanding}. Auditory icons are intuitive associations between a recognizable sound and a piece of information, earcons are unintuitive intentionally designed associations, sonification maps information to variations in sound, and speech conveys information verbally \cite{mcgookin2004understanding}.

We derive three auditory interfaces from the literature:
\begin{enumerate}
    \item \textbf{Earcons:} An association from noises to a piece of information. Earcons have often been studied in the HRI community in multimodal systems \cite{johannsen2004auditory} such as autonomous driving \cite{gang2018don}, adaptive automation of telerobotic control \cite{kaber2006investigation}, and healthcare \cite{rosati2013role}.
    \item \textbf{Phrases:} A truncated version of complete speech.  We create this interface to balance the benefits of condensed information while maintaining some psychological or social aspects of speech. Other studies have also attempted to use some sort of hybrid interface such as spearcons, sped up speech, \cite{walker2006spearcons} or audification, using data as sounds \cite{peres2008auditory}.
    \item \textbf{Sentences:} A complete verbal expression most similar to conversational speech. Most of the literature studying audio in HRI uses this type of verbal dialogue. 
\end{enumerate}


\subsection{Audio in Agriculture}
A survey on nontraditional human robot interactions in agriculture highlights some of the benefits of using speech to improve the usability of technology~\cite{rodriguez2018beyond}. 
Moreover, recent research efforts also determined that robots have not reached a level of design that allows for effective communication of faults by untrained users~\cite{honig2018understanding}.
Literacy has been a major barrier preventing farmers who cannot read written instructions from using robots \cite{rodriguez2018beyond}. 
Many studies attempt to address the issue by surveying farmers and developing audio interfaces to assist under-educated farmers in using technology and communicating with robots~\cite{ ghosh2014krishi}. 
These studies discovered that the complexity of a robot system is one of the main challenges farmers face in adapting technology, as employees lack the necessary skills to operate complicated robots. This conclusion underscores the importance of an intuitive and effective human-centered design when it comes to robot management on an autonomous farm.


\subsection{Communicating Robot Failures Using Audio}
Previous user studies on the failure rate in HRI tasks indicate that direct communication is more important than conversational dialogue to overcome a misperception error and complete a task collectively~\cite{schutte2017robot, fischer2014initiating}. Thus, task-oriented robots should focus on concise utterances rather than lengthy dialogue. This conclusion follows the well-researched principle of least collaborative effort in grounding dialogue, i.e. establishing a mutual understanding \cite{kontogiorgos2020towards,kiesler2005fostering}.

However, more detailed speech improves perceived capability~\cite{cha2015perceived}. Restricting the robot’s vocabulary to simple commands or sounds can negatively impact the robot's usability. Because of the robot's limited speech, the human may underestimate the capability of the robot, thus not utilizing it to its full extent which is called underperception. In overperception, the human overestimates the capability of the robot leading to the human’s expectations not being met, and therefore the human is less willing to work with the robot. When a human misperceives a robot’s capabilities, they misuse it, and their acceptance of the robot decreases which can decrease the success of the collective HRI task~\cite{cha2015perceived}. In general the habitability gap, a mismatch between human expectations and technologies' capabilities, limits speech based human-machine interaction~\cite{moore2017spoken}.

Most literature in communicating HRI failures, including all the sources discussed thus far, deal with embodied robots, either in a video or in a physical setting. 
In our remote operation settings, detailed speech may not improve perceived capability as the robot is not as anthropomorphic.
On the other hand, a remote failure communication system may require more collaborative effort to establish grounding. 
In general, strategies that work for humanoid systems do not transfer over to virtual robotic agents \cite{powers2007comparing}. 
Therefore, we further investigate earcon, phrase and sentence communication systems to better understand how a user perceives such a remote operation system and what collaborative effort is needed in the dialogue.

Recent research efforts have made noteworthy progress in studying HRI in multirobot systems including coordination~\cite{yan2013survey}, user task-switching~\cite{goodrich2005task}, and scalability~\cite{humphrey2007assessing, swamy2020scaled}. Scalability is especially a concern in failure scenarios of multi-robot systems due to the exponential growth of state and action spaces \cite{dahiya2022scalable}. However, the effects of different forms of audio on the user's perception of multi-robot remote supervision failure scenarios has not been well investigated.

\section{Methods}
\label{sec:methods}

\begin{figure}[t]
\vspace{5pt}
\centerline{\includegraphics[width=\columnwidth]{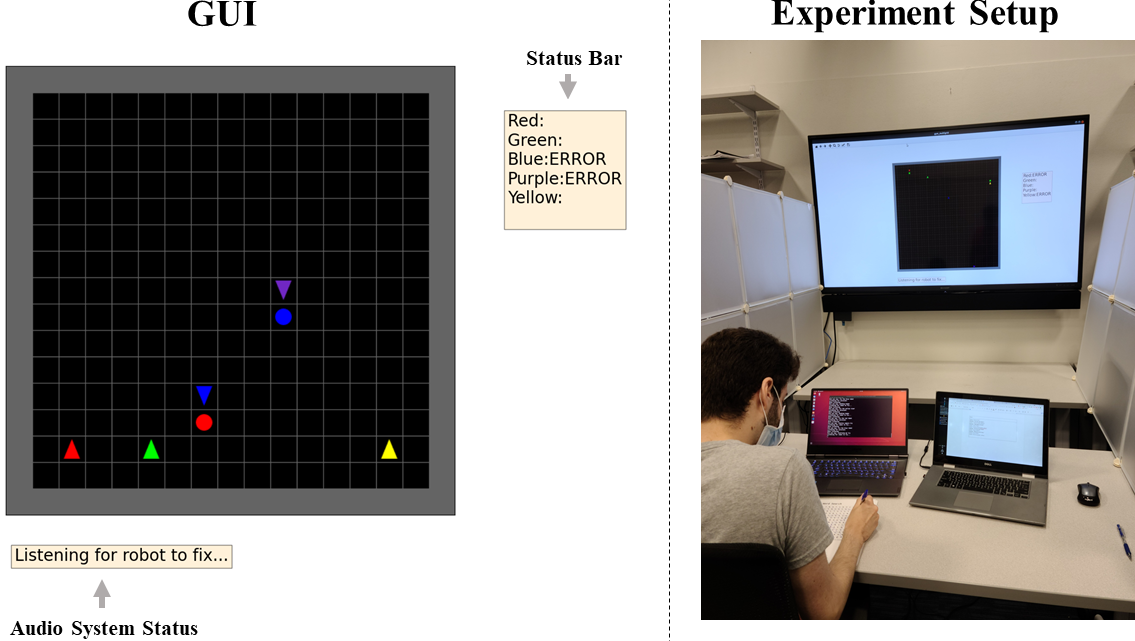}}
\caption{\textbf{\textit{Left}: System GUI}. The status bar indicates which robots have failed (stopped behind circles). Here the purple robot is behind a blue circle (Unrecoverable Failure) and the blue robot is behind a red circle (Row Collision). The red, green and yellow robots are unobstructed and continue to traverse their sections of the grid. The audio system status shows the system is waiting to hear a color indicating which robot to fix.
\textbf{\textit{Right}: Experiment setup.} The setup mimics the control center setting on an autonomous farm. The user is on an isolated desk in front of a TV screen with the GUI and is given wordsearch puzzles to work on. 
}
\label{setup_fig}
\vspace{-10pt} 
\end{figure}

\subsection{Experimental Design}
In order to study the design of an audio interface for agbot fleet monitoring, we develop an autonomous farm environment~\cite{gym_multigrid} as shown in Figure~\ref{setup_fig}. 
Five robots represented as colored triangles navigate up and down columns of the grid sequentially.
Each of the five robots traverses 20\% of the grid and then halts to indicate the completion of the task. 
When a failure, represented as a circle, is reached the robot stops and prompts the user to diagnose the error. 
Based on the error case, the human verbally communicates to the robot on how to resolve the error to reactivate the robot. 

We conduct four experiments in a random order on 13 participants. 
Each experiment prompts the participant differently when a robot fails: earcons, phrases, sentences, and no sound at all. 
Tables~\ref{script_table}-\ref{earcon_table}, are provided to the participant before each experiment. 
After each experiment, we measure the user's perception of success and usability of the system through a survey and their level of productivity through a secondary task score.
The order in which the participant went through the four different conditions was randomized to reduce the bias in the results. 
To prevent unfamiliarity with the system from influencing the results, the participant is provided with a tutorial on how to fix the robots and can practice until they feel comfortable with the system.
The audio system randomly chosen as the fourth experiment is used in the tutorial before the experiments, to discourage participants from considering the tutorial in their survey responses.

\begin{table}[t]
\vspace{5pt}
\caption{Scripts for each audio communication modality
(refer to Table~\ref{earcon_table} for earcon mappings and note how the robot's sentence structure changes in "Sentence" to make it similar to conversational speech)}
\vspace{-10pt}
\begin{center}
\resizebox{\linewidth}{!}{%
\begin{tabular}{p{\dimexpr 0.5\linewidth-2\tabcolsep}|p{\dimexpr 0.5\linewidth-2\tabcolsep}}
\toprule
\textbf{Earcon} & \textbf{Phrase} \\
\midrule

Robot: “[red]”

\textit{Human: “Fix the red robot.”}

Robot: “[robot\_fail]” x 2

\textit{Human: “Navigate around.”}

Robot: “[robot\_fixed]”

Robot: “[blue], [green]”

\textit{Human: “Fix the blue robot.”}

Robot: “[robot\_fail]” x 1 

\textit{Human: “Reverse and retry.”}

Robot: “[robot\_fixed]”

Robot: “[green]”

\dots 

& 

Robot: “Error at red.”

\textit{Human: “Fix the red robot.”}

Robot: “Untraversable obstacle.”

\textit{Human: “Navigate around.”}

Robot: “Error fixed.”

Robot: “Errors at blue, green.”

\textit{Human: “Fix the blue robot.”}

Robot: “Row collision.”

\textit{Human: “Reverse and retry.”}

Robot: “Error fixed.”

Robot: “Errors at green.”

\dots \\
 \midrule
\end{tabular}}
\end{center}

\begin{center}
\resizebox{\linewidth}{!}{%
\begin{tabular}{p{\linewidth}}
\toprule
\textbf{Sentence}\\
\midrule

Robot: “There is an error at the red robot.”

\textit{Human: “Fix the red robot.”}

Robot: “The red robot is facing an untraversable obstacle.”

\textit{Human: “Navigate around.”}

Robot: “The error has been fixed.”

Robot: “There are errors at the following robot blue, green.”

\textit{Human: “Fix the blue robot.”}

Robot: “Row collision has occurred at the blue robot.”

\textit{Human: “Reverse and retry.”}

Robot: “The failure has been fixed.”

Robot: “There are still errors at the green robots.”

\dots\\
\bottomrule
\end{tabular}}
\label{script_table}
\end{center}
\vspace{-10pt} 
\end{table}

\subsection{Failure Cases}
In an attempt to recreate realistic failures from the agricultural domain, we consider three common failure cases: row collision, obstacle, and unrecoverable failure, detailed in Table~\ref{failure_table}.
The above failures can be reliably detected on the field and are assumed to be solvable from the operator’s commands~\cite{ji2020multi,ji2022proactive}.

\begin{table}[t!]
\vspace{5pt}
\caption{Failure types with description and recovery solution.}
\vspace{-10pt}
\begin{center}
\begin{tabular}{ p{0.3\columnwidth} p{0.6\columnwidth} }
\toprule
\textbf{Failure Type} &   \\
\midrule
Row Collision 
& \emph{Description:} The robot deviates from the center line and crashes into crops due to navigation failures.
\newline
\emph{Solution:} Reverse and replan the path that tracks the center line. 
\\
\midrule
Obstacle
& \emph{Description:} The robot encounters obstacles, which obstruct the center line, but still has room around to plan a collision free path.
\newline
\emph{Solution:} Navigate around the obstacle and continue the robot’s original trajectory. 
\\
\midrule
Unrecoverable Failure
& \emph{Description:} The robot is in some failure scenarios where it cannot continue without human intervention, e,g, a fully blocked path.
\newline
\emph{Solution:} Send a human to the field to assist the robot to recover.
\\
\bottomrule
\end{tabular}
\label{failure_table}
\end{center}
\vspace{-10pt} 
\end{table}

\begin{table}[t]
\caption{User commands for the different failure modes}
\vspace{-10pt}
\begin{center}
\resizebox{\linewidth}{!}{%
\begin{tabular}{c c l l}
\toprule
\textbf{Failure \#} & \textbf{GUI Icon} & \textbf{Failure Type}  & \textbf{Solution}\\
\midrule
1 
& o\put(-1,2){\color{red}\circle*{8}}
& Row Collision  
& “reverse and retry”
\\
2
& o\put(-1,2){\color{green}\circle*{8}}
& Obstacle
& “navigate around”
\\
3
& o\put(-1,2){\color{blue}\circle*{8}}
& Unrecoverable Failure
& “sending human”
\\
\bottomrule
\end{tabular}}
\label{solution_table}
\end{center}
\vspace{-10pt} 
\end{table}

The participant addresses each of the failure cases using certain verbal commands shown in Table~\ref{solution_table}. Although the one-to-one mapping from failures to solutions can easily be automated without a human operator, such a system can be extended to a scenario where the human operator must make an informed decision. Even in the current system, the operator chooses the order in which to fix multiple robot failures, which is not a trivial task for humans or planners~\cite{swamy2020scaled}. Nonetheless, knowing how or which robot to address is unrelated to the type of auditory prompt the participant hears, thus the triviality of the system does not discount the merit from a user-centered design perspective.
	
As this study focuses on the impact of audio communication on perception and productivity, the number and type of failures is kept constant.
There are exactly $15$ failures in each simulation, five of each failure type. 
However, the location of each failure is randomly sampled from a uniform distribution across the grid to simulate more realistic failure scenarios. 
A failure is not visible on the grid until a robot reaches it, inhibiting the participant from anticipating a failure.

\subsection{Audio Signals}
The verbal interaction from the human to the robot remains constant throughout all the simulations, but the auditory interaction from the robot to the human will change as described in Tables \ref{script_table}.
When one or more robots fail, the system will prompt the participant with audio indicating the colors of the failed robots. 
The participant will have to verbally say a color to indicate which robot they wish to fix, and the system will indicate which type of failure the robot is facing.
The participant then must say the correct command to fix the system. 
Once fixed, the system will notify the participant that the robot has been fixed and continue with the simulation until another robot fails.

To keep the amount of information conveyed by each audio system constant, we developed a mapping from earcons to robot colors, shown in Table \ref{earcon_table}. To convey what type of failure the robot is at, the system plays a coin noise to indicate the failure number shown in Table~\ref{solution_table}: once for row collision, twice for obstacle, and thrice for unrecoverable failure. Before the earcon experiment, an earcon tutorial program played each of the earcons and their definitions as described in Table \ref{earcon_table}.

\begin{table}[t!]
\vspace{5pt}
\caption{Earcon mappings from sound to meaning
}
\vspace{-10pt}
\begin{center}
\begin{tabular}{c c c c c c}
\toprule
\textbf{Robot} & Red  & Green & Blue & Purple & Yellow \\
\midrule
\textbf{Sound}
& Siren
& Leaves Rustle
& Splash 
& Violin
& Taxi Honk

\\
\bottomrule
\end{tabular}
\end{center}
\begin{center}
\begin{tabular}{c c c }
\toprule
\textbf{Condition} & Robot\_fixed & Robot\_fail \\
\midrule
\textbf{Sound}
& Ready
& Coin
\\
\bottomrule
\end{tabular}
\label{earcon_table}
\end{center}
\vspace{-10pt} 

\end{table}

In the no sound experiment, the participants can only refer to the visual GUI to find out if a failure has occurred, which robot has encountered the failure, and what type of failure the robot is facing. Since the participant had no audio prompt, they would have to occasionally look up from the secondary task to address the failures. The GUI shows all the necessary information conveyed by the audio prompts, as described in Figure~\ref{setup_fig}. 


\subsection{Secondary Task and Productivity}
In an autonomous robot control center setting, the operator is unlikely to fully focus on monitoring the robots. 
Instead, the operator would be completing other tasks and be prompted when a robot needs attention. As a controlled secondary task, a wordsearch puzzle is used in our experiment, which asks a participant to find given words going in various linear directions in a grid of letters. 
The wordsearch puzzle serves as a visual stimulus and cognitive load that the participant can engage with as it does not interfere with the auditory interaction of the system we wish to study.
It is very easy for beginners to learn and does not give an advantage to those with more cultural knowledge or math practice (in contrast to crossword puzzles or math questions). 
Many psychology studies use word search puzzles to study distraction or multitasking~\cite{harrellanalysis, mcmahon2011driven}. 

Dividing the participant's awareness across different senses allows us to measure their productivity when switching attention between the two tasks at hand. 
The participant was put in a constant quiet environment such that the system was their only audio stimuli, as shown in Figure~\ref{setup_fig}.
User studies with physical robots often have auditory background that could affect the user's perception of the system~\cite{martinson2007improving}. 
However, our distraction-free environment isolates the audio's effect on the participant's perception of the system and success in solving the wordsearch puzzles. 

The productivity score of each experiment is the measure of how many words the participant found divided by the total time of the simulation (the simulation stops when all five robots reach the end of their last column of crop).
This words per minute score is robust to small technical inconsistencies or pauses in the system as well as how the participant divides their attention. 
If the system momentarily glitches, then the participant has another second to think and find words. 
If the participant only focuses on finding words, the number of failed robots will increase (and eventually come to a complete standstill), thus increasing the time that the simulation takes to complete, which would give them a low productivity rate. On the other hand, if the participant does not focus on finding words at all and only focuses on the robots, they will find less words and receive a low productivity rate. 
Overall, the system encourages the participant to address both the wordsearch task and the robot failure task at the same time, which strengthens our findings on how audio affects efficiency when faced with a visually intensive task. At the beginning of the study, each participant was made aware of this metric and that they should focus on both the simulation and word search.

\subsection{Participants}
A total of 13 participants voluntarily performed the IRB approved user study, each signing a consent form beforehand. Each of the simulations took approximately eight minutes to run, but with the explanations and tutorial the entire process took around one hour. Every participant was a university affiliate; however, they had varying degrees of previous experience in robotics, with an average experience of 3.4 on a scale of 1 (unfamiliar) to 5 (expert). 

\subsection{Survey Questions}

The survey asked a few background questions before the experiment was conducted. 
After each experiment, the participant was asked to rate the following questions on a 7 point Likert scale:

\noindent \textit{Q1: It was easy to diagnose and fix the errors in this system.}

\noindent \textit{Q2: I was successful at guiding robots passed their failures.}

\noindent \textit{Q3: This system was overwhelming to use.}

The questions were created for the purpose of this design experiment, since most standard HRI questionnaires are designed for physical experiments.
However, our questions relate to the competence dimension of RoSAS \cite{carpinella2017robotic} and the likeability dimension of Godspeed \cite{bartneck2009measurement}, which are two common HRI metrics.

After all audio systems were tested, the participant was asked to rank the notification methods from $1$ to $4$ (best to worse) and to leave feedback on the overall system.

\subsection{Hypotheses}

The following three hypotheses were developed and tested:

\textbf{H1: Any audio interface will facilitate better success, usability, and productivity over a purely visual method.}

In a control center setting, the human will likely be addressing robot failures while operating other systems at once, which in our study is analogous to the word search puzzle. 
The sound notification acts as an interrupt, grabbing the participant’s attention as necessary which allows them to maximize the time they spend on the puzzle. 
With the purely visual interface, the participant must look up at the GUI occasionally to check for errors, which may disrupt their focus more frequently and result in them negatively perceiving the system and performing worse on the secondary task.

During the control experiment with no sound prompts, the participant gets to choose when to look up and address robot failures. 
One could argue that with sound notification a participant's train of thought is interrupted, so looking up on their own might improve their performance. 
However, with sound notification, the participant can fully address a robot's issue without ever looking up at the screen. 
Thus even if they loose their thought process, their eyes may still be on the word search and they still may be able to recognize words while addressing the failures.

\textbf{H2: Single phrase communication provides the best user perceived success, usability, and capability of the system.}

Speech capability improves the perception of social ability in a robot \cite{cha2015perceived}. 
However, the principle of least collaborative effort indicates that minimal effort is socially perceived the best~\cite{kontogiorgos2020towards} in task-oriented HRI dialogue. 
Neither of these findings have been tested on a remote HRI task. We hypothesize that the phrase system provides the most intuitive balance between a socially well perceived yet efficient system. 

\textbf{H3: Single phrase communication will result in the most significant improvement in productivity when a user is completing a secondary task. }

We hypothesize that in the control center scenario the participant will not want to be interrupted from the wordsearch with a lengthy description of the problem nor hold a conversation with the system. However, they may appreciate more easily interpretable feedback than an earcon which they have to map to a robot color. Thus, the single-word communication system will likely provide the best balance between the two extremes.

\section{Results}
\label{sec:results}


The survey results are shown in Table \ref{results_table}. As predicted in H1, the system with no sound performs the worst in every metric. 
As the observations are independent and errors can be assumed to be normally distributed, we ran ANOVA and paired T-tests to verify the results.
Using the type of audio system as the treatment groups, ANOVA was performed on the results of each question, with the following findings: Q1 ($p\ll0.001$), Q2 ($p=0.03$), and Q3 ($p\ll0.001$).
Each result is significant with $\alpha = 0.05$ which supports H1.

\begin{table}[b!]
\caption{Average productivity score (words per minute) and survey responses}
\vspace{-10pt}
\begin{center}
\resizebox{\linewidth}{!}{%
\begin{tabular}{cccccc}
\toprule
\textbf{Audio Type} & \textbf{Prod. Score} & \textbf{Q1}  & \textbf{Q2} & \textbf{Q3} & \textbf{Rank} \\
\midrule
\textbf{Earcon}
& $1.49$ & $5.54$ &	$6.15$ & $2.62$ & $2.23$
\\
\textbf{Phrase}
& $\mathbf{1.73}$ & $\mathbf{6.54}$ & $\mathbf{6.77}$ & $\mathbf{1.85}$ & $\mathbf{1.38}$
\\
\textbf{Sentence}
& $1.56$ & $5.85$ &	$5.85$ & $3.00$ & $2.62$
\\
\textbf{No Sound}
& $1.44$ & $3.85$ & $5.46$ & $4.62$ & $3.77$
\\
\bottomrule
\end{tabular}}
\label{results_table}
\end{center}
\vspace{-10pt} 
\end{table}

\begin{table}[b!]
\caption{P-values of Paired T-tests 
}
\vspace{-10pt}
\begin{center}
\resizebox{0.9\linewidth}{!}{%
\begin{tabular}{ccc}
\toprule
\textbf{Question} & \textbf{Phrase and Earcon} & \textbf{Phrase and Sentence} \\
\midrule
Q1 & $0.003$ & $0.041$
\\
Q2 & $0.013$ & $0.008$
\\
Q3 & $0.013$ & $0.014$
\\
\bottomrule
\end{tabular}}
\label{p_table}
\end{center}
\vspace{-10pt} 
\end{table}

\begin{figure}[t]
\vspace{5pt}
\centerline{\includegraphics[width=.8\columnwidth]{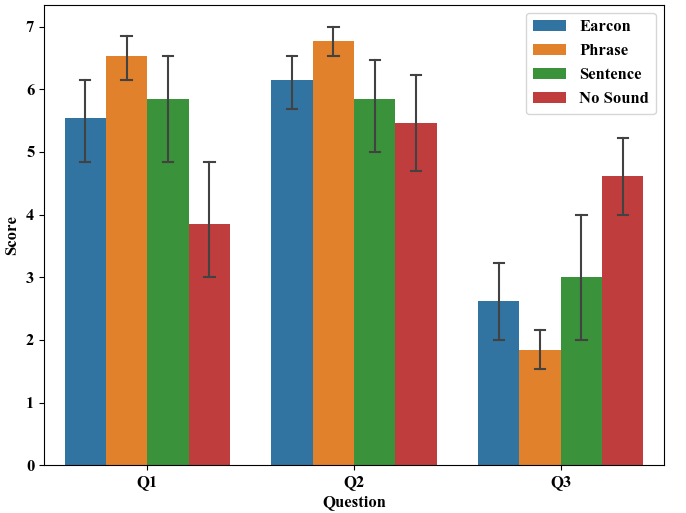}} 
\caption{ Mean survey responses
}
\label{survey_results_fig}
\vspace{-5pt} 
\end{figure}


\begin{figure}[htbp]
\centerline{\includegraphics[width=.8\columnwidth]{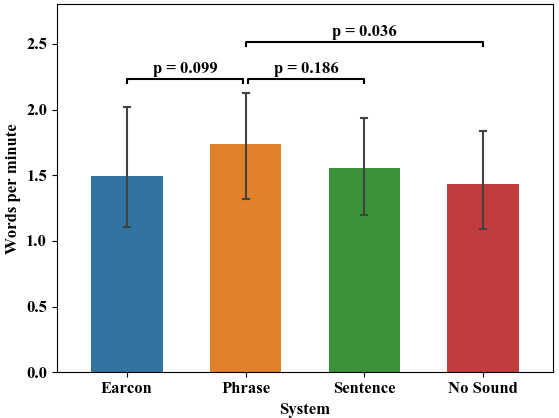}}
\caption{ Mean word search score and p-values of paired T-tests
}
\label{results_fig}
\vspace{-10pt} 
\end{figure}

To verify H2, a visual comparison of the results in Figure~\ref{survey_results_fig} show the mean scores and standard error for each question of each system. 
The survey results show that the user most positively perceives the system prompting them with phrases. 
To test if these results are significant, paired T-tests were performed on each question of the survey results between the phrase system and the two other audio systems. 
These results are statistically significant ($\alpha = 0.05$)  as shown by the p-values in Table~\ref{p_table}. Looking at the rankings provided by the user (Table \ref{results_table}), phrase was ranked better than each of the other methods, which indicates that the user had the most positive impression of the phrase system. H2 is supported by the data.

The results of the secondary task are aggregated in Table \ref{results_table}, and the mean of the phrase system performs slightly higher than the rest. However, ANOVA was performed across the different auditory interaction systems and indicated that the results are not statistically significant ($p=0.78$), thus H3 is not supported. 
We further performed paired T-tests on the productivity scores with the phrase system and every other system and display the significance values in Figure~\ref{results_fig}.

\section{Discussion}
\label{sec:discussion}

\subsection{Explanation of Results}
The data supporting H2 indicates the phrase system most positively affected people's perception of the robots, meaning from a user-centered standpoint it is best suited for completing a human robot teleoperation task, such as fixing failures. 
Phrase notifications were perceived as easiest to use (Q1), most successful (Q2), and least overwhelming (Q3) and was ranked the best.
One reason why sentence performed worse than phrase and earcon could be that the user overperceived the system as capable of more than it actually is since it is closer to full natural language processing.
These unrealistic expectations would increase the chances of an HRI task failure as the human may try to have the system do more than it is capable of, as discovered in \cite{cha2015perceived}. 
For example, in this study the participant often tried to interrupt the system in the middle of the sentence and was forced to repeat when it failed to register the command, making the participant frustrated. 
In general, imagine how many ways a human operator with minimal robotics experience may attempt to use a system when it seems capable of full natural language processing.
Intuitively, it also follows that in a task-oriented remote monitoring setting the robot does not need to have much conversational or social dialogue to be positively perceived by the participant.

Another explanation of why the phrase system scored the best is every other system may have a larger cognitive load on the participant.
The earcon system required the participant to learn the mapping between earcons and robot colors during the experiment. 
Instead of looking at the GUI to see which robot failed, the participant usually would look at the Earcon Table \ref{earcon_table} to determine which robot failed.
After the first few failures, the participant remembered the mappings and did not have to look away from the wordsearch to fix robot failures. 
The sentence system required the largest cognitive load out of the three audio systems.
One participant reported they would “never want a system to speak in full sentences as it is highly frustrating.” 
Communicating one small piece of information took the sentence system a few seconds longer than necessary. 
Such behavior distracted the participant from the wordsearch task and often led the participant to speak before the system was done speaking. 
When the system did not register their response, they had to speak again.
Comprehending a full sentence and being cognizant of when the system finishes the sentence increased the cognitive load on the participant. 
Finally, the no sound system involved the largest cognitive load overall, since the participant always had to remember to look up from the wordsearch to address the failures.
During the audio experiments, the participant rarely took their eyes off the word search, which highlights how audio can aid an HRI interface and reiterates H1.

The participants preferred the earcon system over the full sentence system.
One participant found “it was helpful to have an association between the sounds and the robots that needed fixing.” 
However, all comments about the sentence system given by the participants were negative.
Furthermore, if we remove the worst performing participants (the outliers) from our analysis, the earcon system performs better than the sentence system in all measures.
This indicates that earcon system may be a slightly better design than the sentence system from the user's perspective, however this wasn't statistically analyzed as the phrase methodology was much more positively regarded than both the other methods. 

The larger cognitive load amongst the no sound, earcon, and full sentence conditions is most likely why the single phrase mechanism was the best. This cognitive load may be the effort grounding dialogue literature refers to in the principle of least collaborative effort \cite{kontogiorgos2020towards,kiesler2005fostering}.
Our study extends this principle to remote robot HRI for multitasking, and shows communicating the most information in the least amount of sound is a good design choice. 
However, too little sound like the earcons requires the participant to still do some work to extract the information themselves so the system must take that into account.

\subsection{Limitations}

To our surprise, the devised secondary task and its related metric did not produce significant results, leaving H3 unsupported. 
We hypothesized that the phrase system will improve the user’s capability and increase productivity. 
Although H3 was not statistically supported, on average the phrase system provides a higher productivity (Figure~\ref{results_fig}), which can inform future studies investigating multitasking using audio and visual tasks.
One simple explanation for the lack of significance is the small sample size  that was tested. 
Another explanation is the imperfect speech recognition system occasionally mishearing what the human said.
Such mistakes required the participant to say the command repeatedly until the robot understood them, and was often exacerbated by participants who’s first language was not English.
Having to repeat commands likely decreased the participants productivity enough to negate any significant increase in productivity across the sound systems.
Future studies may consider having participants who are farmers for a more realistic participant group. Additionally, they may use a Wizard of Oz approach to prevent technical imperfections from influencing the results. However, this adjustment is less realistic as misunderstandings are inevitable in modern systems especially in noisy environments. 

\subsection{Broader Impact}
The support for H1 shows audio can make a significant difference in a user's perceived usability and success with the system when multitasking, and H2 proves that concise phrases is better than detailed verbal descriptions or simple sound cues.
Thus, human-robot monitoring interfaces can benefit from audio over visual communication even if it cannot develop fully autonomous conversational AI systems. 
We study the agriculture setting; however, the design can be extended to other applications.
Remote operation is also being used in the healthcare \cite{yang2020keep} and manufacturing \cite{wang2015collaborative}.
Improving usability of robot systems allows robots to become more accessible and widespread.
Enabling more people to manage multiple robots while still being able to focus on other tasks, will improve the overall productivity and quality of society.

\section{Conclusion}
\label{sec:conclusion}

In this paper, we studied how remote supervision of an agricultural robot through speech and audio signals affects a user’s perception of the system and productivity in a secondary task.
Understanding this relationship would allow robotics research to focus on developing systems that match a user’s perception to improve the overall human-robot interaction. 
The results indicate that the average user is most likely to find single phrase notifications of a multirobot remote operation interface the easiest to interact with.

Although productivity was not improved significantly by the phrase system, it did improve participant's perception. 
This result indicates that the participant's perception in this remote robot management scenario had a more noticeable effect on informing the design of the system than the participant's productivity with the system, which underscores the necessity of human-centered design.
We found that participants preferred a system that communicated with simple phrases, followed by simple sound communication. 
Full sentence communication was ranked last by participants.  
These findings match our intuition that full natural language understanding capabilities may not be needed in remote robot monitoring systems as people often found them more annoying than helpful. 

Understanding the optimal interface and communication mechanisms for agricultural robots will help design technology and robots that are easy to use by non-experts. 
What seems intuitive to an academic may not be as easy to use for a person with minimal robotics experience. 
Thus, the applied survey methodology is ideal for alleviating different  experience levels and assuring the accessibility of technology to all types of users.
By improving the design of such fleet management systems, we may increase the likelihood of adoption, paving the way for agbots to be deployed at scale.






\bibliographystyle{IEEEtran}
\bibliography{IEEEabrv,BibFile,agbot-bib}

\end{document}